\documentclass[runningheads]{llncs}
\usepackage[T1]{fontenc}
\usepackage{amsmath}
\usepackage{booktabs}
\usepackage{graphicx}
\usepackage{multirow}
\usepackage{float}
\usepackage{pdflscape}
\usepackage{graphicx}
\begin{document}
\title{Do Encoders Suffice? A Systematic Comparison of Encoder and Decoder Safety Judges for LLM Adversarial Evaluation}
%
%
\author{Han Jeon\inst{1} \and
Shiv Medler\inst{1} \and
Joseph Voyles\inst{1}
\and
Matt Wood\inst{1}
}
%
%
\institute{PricewaterhouseCoopers, U.S.A. \\
\email{Correspondence: han.jeon@pwc.com}\\
\url{}
}
\maketitle              
\begin{abstract}
With the widespread adoption of large language models (LLMs) in chatbots and everyday applications, companies are seeking guardrails that are low on latency and cost but effective. Safety evaluation of LLM outputs has generally relied on LLM-based judges which are effective but slow and expensive at scale. In this paper, we evaluate whether fine-tuned modern encoder classifiers from the ModernBERT family (ModernBERT, Ettin) can reliably identify harmful LLM outputs in user-model conversations without significant performance loss relative to LLM-based judges. We benchmark these encoder classifiers against rules-based prefix matching, fine-tuned LLM classifiers, and LLM judges with varying judge prompt strategies across open-source adversarial datasets. The LLM judges include evaluation methodologies from StrongReject, ShieldGemma, JailbreakBench, AILuminate, SorryBench, a Claude-as-a-judge, and fine-tuned safety classifiers including LlamaGuard 3 and 4. The encoder classifiers are fine-tuned on judge-labeled data under a majority-voting label strategy, then evaluated against a golden holdout dataset to assess how they performed in comparison to the LLM judges. Absolute performance is reported via F1, false negative rate, and precision-recall. We break down results by attack techniques, specifically single-turn prompting, decomposition, escalation, and context manipulation to identify where encoder classifiers align with or diverge from LLM judges. The findings offer guidance on when encoder classifiers can serve as a cost and latency efficient alternative to LLM-based evaluation.

\end{abstract}

\section{Introduction and Related Works}

Large language models (LLMs) are increasingly being deployed in various significant settings including banking chatbots, airline customer service, and other customer-facing applications; however, these LLMs are susceptible to adversarial prompts and jailbreak (Chao et al., 2024). Practitioners must balance guardrail effectiveness, API cost, and latency as these systems grow in complexity and criticality (Wang et al., 2023a). Currently, the dominant paradigm relies heavily on decoder-based safety judges: either large autoregressive models used with LLM-as-a-judge prompts (Zheng et al., 2023) or smaller fine-tuned decoder such as LlamaGuard (Meta AI, 2024). While both are effective, these approaches have the same problems related to cost and latency of inferences, and performance can degrade under distribution shifts (Schwinn et al., 2026) leading to failures to generalize against novel attack patterns unseen during training. While multi-pass strategies have been shown to improve decoder performance on instruction-following (Deng et al., 2023) and reasoning (Wang et al., 2023b), they further highlight that decoders do not natively process context with the completeness and understanding that bidirectional encoders provide by design.

Safety classification is fundamentally a language understanding task since determining whether a response is harmful requires reading and comprehending the full input rather than generating new tokens. We argue this makes discriminative encoder architectures a natural fit because encoders process the entire input bidirectionally in a single pass, capturing harmful intent across the full context without the sequential left-to-right constraints of autoregressive decoding. Recent work confirms encoders outperform larger decoder-only models on classification tasks (Weller et al., 2026), and ModernBERT (Warner et al., 2024) has renewed interest in this architecture as it comes with extended context windows and modern pretraining. Thus we look at how the family of ModernBERT models perform in adversarial safety evaluation which have ambiguous patterns and guidelines from both questions and answers. This paper addresses that gap by fine-tuning a ModernBERT-family encoder (Ettin) on labels aggregated from seven safety judges via majority vote, then evaluating the resulting classifier on a held-out benchmark against four frontier model baselines.

Our contributions consist of:

\begin{itemize}
\item The first systematic comparison of modern encoder classifiers against both LLM-as-a-judge systems and fine-tuned decoder-based safety models on adversarial safety benchmarks, using ground-truth holdout labels.
\item A cost-latency-performance analysis in production systems quantifying the practical trade-offs of encoder classifiers as guardrails relative to decoder-based alternatives.
\item A fine-grained breakdown of classifier performance by attack technique that includes single-turn prompting, decomposition, escalation, and context manipulation. This identifies where encoder classifiers align with or diverge from decoder-based judges.
\end{itemize}

\subsection{LLM-as-a-Judge and Decoder-Based Safety Evaluation}

Recent safety evaluation work broadly falls into two categories: integrated benchmark frameworks and standalone datasets. Benchmarks such as StrongReject (Souly et al., 2024), SorryBench (Xie et al., 2025), and HarmBench (Mazeika et al., 2024) combine prompt sets with built-in evaluation pipelines, while datasets like AdvBench (Zou et al., 2023), ToxicChat (Lin et al., 2023), BeaverTails (Ji et al., 2023), and XSTest (R\"ottger et al., 2023) provide adversarial prompts and annotations without a fixed evaluation protocol.

Across both categories, the LLM-as-a-judge paradigm (Zheng et al., 2023) has emerged as the default scoring mechanism: flexible and context-sensitive, but expensive at scale. In this work, we use judge methodologies from StrongReject, SorryBench, and HarmBench as training signals, and evaluate against ground-truth holdout labels from JailbreakBench and AILuminate.

Another approach to the safety classification problem is to use fine-tuned decoder classifiers. There are several variants including LlamaGuard (Meta AI, 2024), Qwen3Guard (Alibaba, 2024), and ShieldGemma (Google, 2024) that are autoregressive models that produce safety labels based on defined harm taxonomies. These models benefit from instructions during fine tuning and learned to handle multi-turn conversational context at the cost of being multi-billion parameters. This causes significant inference latency and requires repeated fine-tuning as attack styles evolve. In our evaluation, we leverage LlamaGuard 3 and 4 as training-signal judges and baselines in our results.

\subsection{Encoder-Based Safety Classification}

Prior encoder work applied Sentence-BERT (Zheng et al., 2024) and BERT-based classifiers (Kumar et al., 2025) to safety tasks, but both are before modern encoder architectures came into the scene. With the release of ModernBERT (Warner et al., 2024), there was renewed interest in encoders for classification with techniques like PangolinGuard (Carpintero 2024), which used it for prompt injection detection, and Soares et al. (2025) which explored contrastive training for encoder robustness. Broader encoder-decoder comparisons exist in adjacent domains: Ettin (Weller et al., 2026) provides a new state-of-the-art encoder in the ModernBERT family and RedLLM (Zhang et al., 2025) examines scaling behavior across model classes, but neither targets safety evaluation, interest in encoders for classification with techniques like PangolinGuard (Carpintero 2024), which used it for prompt injection detection, and Soares et al. (2025) which explored contrastive training for encoder robustness. Broader encoder-decoder comparisons exist in adjacent domains: Ettin (Weller et al., 2026) provides a new state-of-the-art encoder in the ModernBERT family and RedLLM (Zhang et al., 2025) examines scaling behavior across model classes, but neither targets safety evaluation. No prior work systematically benchmarks modern encoder classifiers against decoder-based judges for adversarial safety classification.

\section{Dataset}

\subsection{Dataset Construction - Source Datasets and Domain Coverage}

The training corpus aggregates 14 source datasets spanning standardized safety benchmarks, real-world interactions, adversarial jailbreak collections, and domain-specific synthetic data. Of these, 11 are external datasets and 3 are synthetically generated (medical, financial, internal). Standardized safety benchmarks include HarmBench (Mazeika et al., 2024), SafetyBench (Zhang et al., 2023), and Simple Safety Tests (Vidgen et al., 2024). HarmBench and Simple Safety Tests provide harmful behaviors and prompts for evaluating refusal performance. SafetyBench differs in format: it consists of multiple-choice questions designed to assess safety understanding rather than adversarial robustness. As such, it is used as a complementary signal rather than being directly integrated into the adversarial prompting pipeline.

Real-world and human-annotated data include ToxicChat (Lin et al., 2023) and WildJailbreak (Zhang et al., 2024). ToxicChat captures toxicity from real user--AI interactions, while WildJailbreak provides large-scale compositional jailbreak prompts mined from production conversations. We use both harmful and benign splits from WildJailbreak to capture contrastive behavior. Adversarial and evaluation-focused datasets include AdvBench (Zou et al., 2023), OR-Bench Hard and OR-Bench Toxic (Cui et al., 2024), BeaverTails-Eval (Ji et al., 2023), Do-Not-Answer (Wang et al., 2023c), and XSTest (R\"ottger et al., 2023). These datasets provide a mix of direct harmful instructions, over-refusal probes (OR-Bench Hard), genuinely harmful prompts (OR-Bench Toxic), and boundary-testing examples (XSTest). BeaverTails-Eval refers specifically to the curated 700-prompt evaluation subset with human annotations, rather than the full dataset.

Domain-specific synthetic datasets are constructed for medical and financial safety. For each domain, harm categories are defined, and OpenAI GPT-5.2 is used to generate harmful prompts paired with chain-of-thought explanations. All generated samples are manually reviewed to ensure each prompt represents a distinct attack. An additional internal synthetic adversarial dataset (150 prompts) is included and created in the same way. Each source prompt is executed across a combination of target LLMs and adversarial prompting techniques, producing multiple (prompt, response) pairs per input. Primary datasets, which provide broad adversarial diversity or cover unique domains, receive full inference coverage across all configurations. These include WildJailbreak, SafetyBench, ToxicChat, HarmBench, Internal Adversarial, Medical Synthetic, Financial Synthetic, and Simple Safety Tests, totaling 5{,}406 source prompts and 537{,}477 inference rows. Supplementary datasets, which include AdvBench, OR-Bench Hard, OR-Bench Toxic, BeaverTails-Eval, Do-Not-Answer, and XSTest, increase category breadth but are subsampled due to overlap with primary sources. They contribute 4{,}562 prompts and 25{,}739 rows. In total, the training corpus consists of 9{,}968 source prompts and 563{,}216 (prompt, response) pairs.

\noindent\footnotesize{* Contains both safe and harmful prompts.}\normalsize

\subsection{Leakage Prevention and Dataset Partitioning}

Each of the 16 target models in our inference pipeline produces a response to every question in the corpus, meaning the same adversarial question appears up to 16 times the dataset, once per model, and further across different attack techniques and configurations. A naïve row-level random split would be insufficient, as model A’s response to a question in the training set would potentially be a precursor to also fooling model B’s response to the same question in the test set. Because the classifier can learn question-level features, this introduces data leakage from training to evaluation. The model is performing a memorization task rather than a true response-level harmfulness detection.

To mitigate this, we apply a question group partitioning strategy. All rows sharing the same question text, regardless of model, attack technique, or generation configuration, are assigned to the same partition prior to any stratification. Splits are constructed using \texttt{StratifiedGroupKFold} with $\text{groups} = \text{question\_id}$ and stratification based on the majority-vote harmfulness label. This ensures that the class distribution (harmful vs. safe) is preserved across k-folds while maintaining question-level splits between training and test data. The resulting configuration uses $5$-fold cross-validation for model training.

\subsection{Holdout and Out-of-Distribution Evaluation Data}
A hold-out evaluation set of 42{,}899 examples is constructed from two external benchmarks: JailbreakBench Behaviors (Chao et al., 2024), consisting of 200 behaviors, and AILuminate v1 Demo (industry benchmark, 2024), consisting of 1{,}200 test cases with ground-truth labels, for a total of 1{,}400 source items. The ground-truth labels are inherited directly from JailbreakBench and AILuminate and correspond to prompt-level labels rather than response-level adjudications. Thus, JailbreakBench labels indicate the safety status of the source behavior, while AILuminate labels indicate the benchmark-defined safety category of each prompt. No additional response-level labeling is performed on the generated model outputs.

Each source item is evaluated across four frontier models (Claude Sonnet 4.5, Gemini 3.1 Pro, GPT-5.4, and GPT-OSS 120B), four adversarial attack techniques (single-turn, decomposition, escalation, and context manipulation), and two temperature settings, yielding 44{,}800 examples. The inference success rate is 95.5\%, with 1{,}901 failures removed from the original 44{,}800 generations, resulting in a final set of 42{,}899 examples. No examples from these benchmarks are included in the training pool at any stage. This held-out set is used exclusively for out-of-distribution (OOD) evaluation after all model selection decisions have been finalized on the validation split.

\section{Adversarial Attack Methodology}

\subsection{Target Model Selection}
                                              
We evaluate adversarial robustness across 16 LLMs from six providers (OpenAI, Anthropic, Meta, Mistral, Google, and Amazon) using a fully automated pipeline that produces (prompt, response) pairs that are used for subsequent evaluation. All models are queried via API without custom system prompts at temperatures $T = 0.0$ and $T = 0.7$. $T = 0.0$ establishes a greedy-decoding baseline with $T = 0.7$ tests whether some randomness when picking tokens will lead to harmful results. o4-mini does not expose a temperature parameter and is queried once per prompt. The model sets are organized into two generational cohorts. The legacy cohort (pre-2025) includes GPT-4o-mini, Claude 3 Haiku, LLaMA 3/3.3 (70B), Mixtral 8x7B, Mistral Large, and Amazon Nova Pro; the modern cohort (2025 onward) includes GPT-4.1, o4-mini, Claude Sonnet/Haiku 4.5, LLaMA 4 Maverick, Pixtral Large, and Gemini 2.5/3/3.1. The diversity of having six providers and two generations serves two purposes: add variations of refusal styles from multiple vendors, and newer generations produce harmful outputs that are significantly different from older models. By querying all 16 models per prompt, there is a guarantee of sufficient variation of responses per question which allows for uniqueness when grouped together in training as seen in §2.3. Table~\ref{tab:model_table} has the full details on the models groupings.

\begin{table}[t]
\caption{Target LLMs used in evaluation}
\label{tab:model_table}
\centering
\resizebox{\textwidth}{!}{
\begin{tabular}{l l l l l l l}
\hline
\# & Model & Provider & Cohort & Release & Input (\$/1M) & Output (\$/1M) \\
\hline
1 & GPT-4o-mini & OpenAI & Legacy & Jul 2024 & 0.15 & 0.60 \\
2 & Claude 3 Haiku & Anthropic & Legacy & Mar 2024 & 0.25 & 1.25 \\
3 & LLaMA 3 70B Instruct & Meta & Legacy & Apr 2024 & 0.99 & 0.99 \\
4 & LLaMA 3.3 70B Instruct & Meta & Legacy & Dec 2024 & 0.72 & 0.72 \\
5 & Mixtral 8x7B Instruct & Mistral & Legacy & Dec 2023 & 0.45 & 0.70 \\
6 & Mistral Large 2407 & Mistral & Legacy & Jul 2024 & 2.00 & 6.00 \\
7 & Amazon Nova Pro v1 & Amazon & Legacy & Dec 2024 & 0.80 & 3.20 \\
8 & GPT-4.1 & OpenAI & Modern & Apr 2025 & 2.00 & 8.00 \\
9 & o4-mini & OpenAI & Modern & Apr 2025 & 1.10 & 4.40 \\
10 & Claude Sonnet 4.5 & Anthropic & Modern & Sep 2025 & 3.00 & 15.00 \\
11 & Claude Haiku 4.5 & Anthropic & Modern & Oct 2025 & 1.00 & 5.00 \\
12 & LLaMA 4 Maverick 17B & Meta & Modern & Apr 2025 & 0.24 & 0.97 \\
13 & Pixtral Large 2502 & Mistral & Modern & Feb 2025 & 2.00 & 6.00 \\
14 & Gemini 2.5 Flash & Google & Modern & May 2025 & 0.30 & 2.50 \\
15 & Gemini 3 Flash (preview) & Google & Modern & Dec 2025 & 0.50 & 3.00 \\
16 & Gemini 3.1 Pro (preview) & Google & Modern & Feb 2026 & 2.00 & 12.00 \\
\hline
\end{tabular}
}
\end{table}

\subsection{Attack Techniques}

Each model is evaluated under four adversarial prompting techniques: single-turn pass-through, decomposition, escalation, and context manipulation. As Table 2 shows, large-scale adversarial experiments are expensive per API costs as these costs are multiplied across model, technique, and temperature settings. Evaluation compounds this as frontier judges carry expensive API costs. The single-turn baseline sends one adversarial prompt directly to the target model and records the response, which provides a clean reference point against which multi-turn techniques are compared. Prompts are drawn from AdvBench (Zou et al., 2023) and STRONGReject (Souly et al., 2024). The three multi-turn strategies share a common structure: an attacker LLM (Claude Haiku) constructs the full multi-turn dialogue first, then passes it to the target model for completion. The attacker never interacts with the target in real time; the entire conversational trajectory is planned.

\textbf{Decomposition} (6 API calls) fragments a harmful request into four sub-queries, each benign in isolation. The target answers each independently without access to the overall intent, and a final synthesis step prompts the model to combine the prior answers into a single unified output, reconstructing the original harmful objective (Zhou et al., 2024).

\textbf{Escalation} (4 API calls) begins with benign rapport-building turns and incrementally increases the sensitivity of each subsequent request. The target model responds at each step with access to the full conversation history, so earlier cooperative responses become part of the context window when the harmful request arrives (Russinovich et al., 2024).

\textbf{Context manipulation} (3 API calls) embeds the harmful request inside a benign framing sequence which includes academic analysis, fictional narrative, or professional scenario, so that it reads as a natural continuation of an established context rather than a standalone harmful query (Yu et al., 2024).

\section{Safety Judge Panel}

We assemble a panel of seven judges across three architectural categories: open-weight classifiers hosted on private infrastructure (LlamaGuard 3, LlamaGuard 4), proprietary LLMs prompted for safety classification (Claude Judge and SorryBench, JailbreakBench Judge, AILuminate Judge), and a rubric-based continuous scorer (StrongReject). Each prompted judge runs at temperature 0.0 on a dedicated backbone: Claude Judge and StrongReject use Claude Haiku 4.5, SorryBench uses GPT-4o-mini, JailbreakBench Judge uses LLaMA 3.3 70B, and AILuminate Judge uses Claude Sonnet 4.5. Two of these, JailbreakBench and AILuminate, replicate the evaluation protocols of their respective public leaderboards, enabling direct comparison against published attack success rates. An eighth heuristic judge (Prefix Match) is retained as control and for descriptive analysis but excluded from label aggregation. No single judge is treated as ground truth. Individual judges carry distinct calibration biases and failure modes, which are averaged out through a majority vote. Inter-judge agreement statistics (Fleiss' $\kappa$) are the main object of analysis and feed directly into the high-confidence filtering described below. Binary labels are produced by the majority vote of the seven judges with each judge providing a single binary label of harmful or safe per response. Any prompt-level annotations, partial-coverage, and intermediate scores are excluded to ensure that the row reflects a clean consensus. Any judges that fail to return a valid label are removed from the numerator and denominator, acting as an abstain vote, which may occur from endpoint failure.  
\[
\frac{\sum \text{harmful\_votes}}{\sum \text{responding\_judges}} > 0.5
\]

This introduces row-level variation in the effective voting threshold.

\paragraph{High-Confidence Filtering.}
Fleiss' $\kappa$ across all seven judges on the full dataset is approximately 0.23. This is low, but the disagreement is structured in which compliance-oriented judges (StrongReject) and taxonomy-based classifiers (LlamaGuard) diverge systematically rather than randomly. To isolate rows where judges substantively agree, we retain only unanimously safe rows (0 of $N$ non-null judges vote harmful) and strong-majority harmful rows ($\geq 5$ of 7), discarding the 1--4 vote region entirely. This cuts the dataset from approximately 563K to 307K rows — a 45.4\% reduction — and raises $\kappa$ to approximately 0.805 in the train. This filtering improves label quality at the expense of data coverage by removing ambiguous answers where answers are contested. Classifiers are trained under both the full and high-confidence conditions as a controlled ablation: if the filtered variant produces higher held-out precision and recall, annotation noise is the bottleneck, not model capacity.

\paragraph{Training signal analysis.}
Inter-judge agreement is analyzed using pairwise Cohen's $\kappa$, Fleiss' $\kappa$, and complementary statistical tests to assess divergence, label consistency, and systematic differences in decision making. There is substantial disagreement in the full training dataset (Fleiss' $\kappa = 0.232$), which improves to ($\kappa = 0.805$) on the high-confidence subset, indicating effective filtering. Cohen's pairwise scores show strong disagreements between judges with StrongReject and Claude ($\kappa = 0.033$) and StrongReject and LlamaGuard (LG4: 0.102, LG3: 0.115), while LLM-based evaluators such as SorryBench and AILuminate are more aligned: ($\kappa = 0.668$). Agreement is the lowest for decomposition attacks ($\kappa = 0.095$), highlighting almost random multi-turn ambiguity. These differences are statistically significant (Cochran’s $Q = 295{,}589$, $p \approx 0$), and all judges exhibit category-dependent biases ($\chi^2$, $p \approx 0$). These findings were the reason for the high-confidence filtering strategy.

\section{Evaluation Protocol}

Our encoder classifiers are trained on majority-vote labels aggregated from seven judges with varying design philosophies, meaning the training signal is a blend of different notions of harmfulness. The evaluation protocol is designed to first characterize the structure of that blended signal then measure how effectively classifiers learn from it.

\paragraph{Classifier performance.}
We evaluate on an OOD holdout set (JailbreakBench + AILuminate; 42,899 rows across 4 frontier models). The holdout set carries ground-truth labels from the original benchmark authors, so evaluation measures classifier accuracy against human-adjudicated annotations rather than against the majority-vote signal used for training. The holdout tests generalize across three dimensions simultaneously: dataset shift, since neither benchmark appears in training; model shift, since the holdout models include frontier systems released after the training pool was assembled; and evaluation alignment, since judges follow official benchmark protocols, enabling direct comparison with published attack success rates.

\paragraph{Metrics and threshold selection.}
F1 is the primary metric; false negative rate (FNR) is reported separately as the safety-critical measure, since a missed harmful response carries asymmetric cost. AUROC provide threshold-independent assessment; however, it is not reported in the main results table because it is not available for all evaluated models. Encoder models achieve high AUROC: Ettin-150M reaches 0.939 and Ettin-400M reaches 0.929, indicating strong separation between harmful and safe responses. The decision threshold is optimized over $[0.1, 0.9]$ on the validation set to avoid extremes where classifiers trivially predict all or no outputs as harmful. All classifiers are evaluated on the holdout using these validation-selected thresholds. Comparing performance across the full and high-confidence training conditions isolates whether label noise from the blended signal, rather than model capacity, is the binding constraint on downstream accuracy. All metrics include 95\% bootstrap confidence intervals (1{,}000 resamples).

\section{Results}

\subsection{Encoder Performance on the OOD Holdout}

Encoder classifiers show two concrete advantages over decoder-based judges on the OOD holdout dataset. First, they exhibit more stable calibration behavior. LlamaGuard-4 under-predicts harmful content, with a harmful rate of 3.5\% and FNR of 0.658, while Claude Judge and StrongReject over-predict, each with harmful rates of 25.1\% and recall values of 0.994 and 0.900, respectively. In contrast, the encoder models maintain more balanced operating characteristics: Ettin-150M has a harmful rate of 4.79\% and FNR of 0.422, while Ettin-400M reduces FNR to 0.146 at a harmful rate of 13.44\%.

Second, the two encoders offer complementary operating points. Ettin-150M provides a balanced profile (F1 = 0.547, precision = 0.519, recall = 0.578, FNR = 0.422), while Ettin-400M operates as a high-recall detector (recall = 0.854, FNR = 0.146) with lower precision (0.304) and higher harmful rate (13.44

To assess the impact of label uncertainty, we additionally evaluate a high-confidence subset derived from the same OOD holdout dataset by retaining only examples with strong inter-judge agreement ($\kappa \approx 0.81$). On this filtered OOD subset, Ettin-150M achieves an F1 score of 0.800 and Ettin-400M achieves an F1 score of 0.894. The substantial improvement relative to the full OOD holdout suggests that annotation noise, rather than model architecture, is a primary limiting factor. Full results are provided in Appendix A.

\subsection{Failure Modes Under Multi-Turn Attacks}

Encoder-based classifiers consistently struggle with multi-turn attacks, particularly decomposition, where harmful intent is distributed across intermediate steps and only becomes apparent when aggregated across turns. Because the classifier observes only the final (question, response) pair, it never sees the intermediate interactions in which harmful intent is fragmented across otherwise benign sub-queries. Although the encoder models process the full context window available in the final exchange, they lack direct observability into the reasoning path that produced the response. This limitation is reflected in the results, where decomposition attacks produce the highest false negative rates among the evaluated techniques. Increasing model size partially mitigates the issue: the 400M encoder reduces the overall false negative rate from 42.2\% to 14.6\%, and more notably from 50.6\% to 11.6\% on decomposition attacks. These results suggest that larger encoder models are more robust to distributed harm patterns even without explicit access to intermediate turns.

While the 150M model achieves higher overall F1, the 400M model consistently achieves lower false negative rates across all attack types. The most significant improvement occurs for decomposition attacks, where the model transitions from missing roughly half of harmful responses to detecting nearly 90\%. Nevertheless, this remains a fundamental limitation of encoder-based approaches. Modern jailbreak techniques increasingly rely on multi-turn strategies such as prompt decomposition, escalation, and context manipulation, where harmful intent emerges only through the accumulation of context across interactions. Because encoder classifiers operate on a fixed classification input rather than the full conversation, they may fail to accurately assess sophisticated attacks that require reasoning over conversation turns or user intent. As a result, encoder-based classifiers are best viewed as efficient first-line filters, while higher-cost decoder-based judges or trajectory-aware systems may still be necessary for comprehensive evaluation of advanced multi-turn jailbreaks. Full evaluation results by technique are provided in Appendix A.

\subsection{Practical Implications: Cost and Latency Trade-offs}

Differences between encoder- and decoder-based approaches become particularly apparent in production environments where safety filtering must operate continuously and at scale. Ettin-150, Ettin-400, and LlamaGuard-4-12B were benchmarked on an Azure ML \texttt{Standard\_ND40rs\_v2} instance with 8 NVIDIA Tesla V100 GPUs using 100 conversations from the Anthropic Red Team Attempts dataset. All models were evaluated in FP16 at batch sizes of 1, 8, 32, and 64. Throughput was measured in completions per second (CPS), and latency using P50, P95, and P99 response times.

Results are reported in Appendix~B. LlamaGuard-4-12B achieved 1.10--1.31 CPS with median latencies of 718--917 ms, whereas the Ettin models achieved 50.2--244.8 CPS with median latencies below 20 ms. At batch size 8, Ettin-150 achieved 244.8 CPS versus 1.31 CPS for LlamaGuard-4-12B, a 187$\times$ throughput improvement while reducing median latency from 717.6 ms to 3.6 ms. On the same Azure ML \texttt{Standard\_ND40rs\_v2} instance (8 NVIDIA Tesla V100 GPUs; \$22.03/hr) (Azure, 2026), Ettin-150 achieved 11.1 CPS per dollar per hour compared to 0.06 for LlamaGuard-4-12B. Furthermore, Ettin models can run on low-cost Azure ML CPU instances (\texttt{Standard\_E4s\_v3}, \$0.25/hr) (Azure 2026), further reducing deployment costs. These results highlight the efficiency advantages of encoder-based approaches for real-time safety filtering. Azure ML pricing was obtained from Microsoft Azure pricing documentation at the time of evaluation.

\section{Conclusion}

We present the first large-scale evaluation of fine-tuned encoder classifiers as safety judges, which are benchmarked against decoder-based approaches in 563,216 q-a pairs encompassing 16 target models, 14 datasets, and four variations of attack techniques. The finetuned Ettin classifier outperformed several safety judges that were significantly larger (e.g., LlamaGuard-3/4, Claude Judge, StrongReject) on the OOD holdout set, while remaining below the strongest benchmark-aligned evaluators (AILuminate, SorryBench). These results indicate that encoders can be used as a first-pass filter to safety classification before passing on more difficult problems to decoder-based models. From a cost and latency perspective, encoder classifiers are significantly cheaper and more effective than LlamaGuard 3 and 4 and some LLM-as-a-judge strategies. Future work will extend to testing on human curated datasets as inter-judge agreement was fairly low (Fleiss' $\kappa$ = 0.232) confirming noisy labels during training and constrained the true performance of encoder-based classifiers.



\clearpage
\appendix

\section{Evaluation Results}

\vspace{4pt}

\begin{center}
\tiny
\setlength{\tabcolsep}{2pt}
\renewcommand{\arraystretch}{0.85}

\textbf{F1 and FNR by Technique}\\[2pt]

\begin{tabular}{lccccc|ccccc}
\toprule
&
\multicolumn{5}{c|}{F1} &
\multicolumn{5}{c}{FNR} \\
\cmidrule(lr){2-6}
\cmidrule(lr){7-11}
Judge &
All & Ctx & Dec & Esc & Single &
All & Ctx & Dec & Esc & Single \\
\midrule
AILuminate   & 0.767 & 0.795 & 0.764 & 0.729 & 0.729 & 0.208 & 0.183 & 0.149 & 0.332 & 0.354 \\
SorryBench   & 0.706 & 0.785 & 0.699 & 0.608 & 0.642 & 0.112 & 0.081 & 0.148 & 0.090 & 0.093 \\
JBB          & 0.615 & 0.657 & 0.551 & 0.661 & 0.656 & 0.493 & 0.446 & 0.573 & 0.387 & 0.437 \\
Ettin-150M   & 0.547 & 0.604 & 0.505 & 0.499 & 0.572 & 0.422 & 0.393 & 0.506 & 0.332 & 0.305 \\
Ettin-400M   & 0.449 & 0.511 & 0.461 & 0.330 & 0.402 & 0.146 & 0.183 & 0.116 & 0.176 & 0.132 \\
LG4          & 0.385 & 0.420 & 0.367 & 0.297 & 0.419 & 0.658 & 0.632 & 0.671 & 0.748 & 0.618 \\
Claude       & 0.319 & 0.386 & 0.250 & 0.384 & 0.455 & 0.006 & 0.004 & 0.000 & 0.010 & 0.026 \\
LG3          & 0.309 & 0.380 & 0.288 & 0.238 & 0.291 & 0.553 & 0.531 & 0.545 & 0.618 & 0.583 \\
StrongReject & 0.289 & 0.349 & 0.360 & 0.167 & 0.210 & 0.100 & 0.081 & 0.155 & 0.020 & 0.036 \\
PrefixMatch  & 0.116 & 0.138 & 0.187 & 0.050 & 0.072 & 0.037 & 0.082 & 0.007 & 0.030 & 0.026 \\
\bottomrule
\end{tabular}
\end{center}

\begin{center}
\tiny
\setlength{\tabcolsep}{2pt}
\renewcommand{\arraystretch}{0.82}

\textbf{Full Evaluation Results}\\[-1pt]

\begin{tabular}{lccccc}
\toprule
Judge & F1 & Prec. & Rec. & FNR & Harm. \\
\midrule
AILuminate & 0.767 & 0.743 & 0.792 & 0.208 & 5.1\% \\
SorryBench & 0.706 & 0.585 & 0.888 & 0.112 & 7.3\% \\
JBB & 0.615 & 0.781 & 0.507 & 0.493 & 3.1\% \\
Ettin-150M & 0.547 & 0.519 & 0.578 & 0.422 & 4.8\% \\
Ettin-400M & 0.449 & 0.304 & 0.854 & 0.146 & 13.4\% \\
LG4 & 0.385 & 0.440 & 0.342 & 0.658 & 3.5\% \\
Claude & 0.319 & 0.190 & 0.994 & 0.006 & 25.1\% \\
LG3 & 0.309 & 0.236 & 0.447 & 0.553 & 9.1\% \\
StrongReject & 0.289 & 0.172 & 0.900 & 0.100 & 25.1\% \\
ShieldGemma-9B & 0.154 & 0.090 & 0.524 & 0.476 & 27.9\% \\
PrefixMatch (baseline) & 0.116 & 0.062 & 0.963 & 0.037 & 74.9\% \\
ShieldGemma-2B & 0.053 & 0.063 & 0.046 & 0.954 & 3.5\% \\
\bottomrule
\end{tabular}

\vspace{2pt}

\textbf{High-Confidence (HC) Subset Results}\\[-1pt]

\setlength{\tabcolsep}{1.8pt}
\begin{tabular}{lccccc}
\toprule
Judge & F1 / 95\% CI & Prec. & Rec. & AUPRC / 95\% CI & AUROC \\
\midrule
Claude & 0.999$^\star$ & 1.000 & 0.997 & 0.997$^\star$ & 0.999 \\
SorryBench & 0.982$^\star$ & 1.000 & 0.965 & 0.966$^\star$ & 0.982 \\
AILuminate & 0.968$^\star$ & 1.000 & 0.937 & 0.940$^\star$ & 0.969 \\
StrongReject & 0.964$^\star$ & 1.000 & 0.931 & 0.963$^\star$ & 0.997 \\
Ettin-400M HQ & 0.900 {\tiny(0.887,0.912)} & 0.904 & 0.896 & 0.922 {\tiny(0.901,0.939)} & 0.984 \\
Ettin-150M HQ & 0.853 {\tiny(0.834,0.868)} & 0.900 & 0.810 & 0.883 {\tiny(0.858,0.906)} & 0.978 \\
Ettin-400M NF & 0.851 {\tiny(0.832,0.869)} & 0.984 & 0.750 & 0.960 {\tiny(0.952,0.967)} & 0.993 \\
JBB & 0.835$^\star$ & 1.000 & 0.716 & 0.728$^\star$ & 0.858 \\
Ettin-150M NF & 0.800 {\tiny(0.781,0.819)} & 0.956 & 0.688 & 0.930 {\tiny(0.916,0.940)} & 0.988 \\
LG3 & 0.752$^\star$ & 1.000 & 0.603 & 0.619$^\star$ & 0.801 \\
LG4 & 0.499$^\star$ & 1.000 & 0.332 & 0.360$^\star$ & 0.666 \\
ShieldGemma-9B & 0.163 {\tiny(0.151,0.175)} & 0.098 & 0.486 & 0.089 {\tiny(0.082,0.099)} & 0.737 \\
ShieldGemma-2B & 0.152 {\tiny(0.135,0.167)} & 0.108 & 0.256 & 0.084 {\tiny(0.075,0.092)} & 0.675 \\
\bottomrule
\end{tabular}

\vspace{1pt}
\raggedright
\tiny
\textit{Note.} 
 Confidence intervals are omitted for binary judges; the high-confidence subset is restricted to verified judge-positive examples, yielding precision of 1.000 by construction.
\end{center}

\vspace{2pt}

\section{Inference Throughput and Latency}

\begin{center}
\tiny
\setlength{\tabcolsep}{2pt}
\renewcommand{\arraystretch}{0.82}

\textbf{Inference Performance Across Models and Batch Sizes}\\[-1pt]

\begin{tabular}{lcccccc}
\toprule
Model & Batch & CPS & CPS SD & P50 & P95 & P99 \\
\midrule
\multirow{4}{*}{Ettin-150M}
& 1  & 63.26  & 0.71 & 15.69 & 17.07 & 17.50 \\
& 8  & 244.76 & 4.94 & 3.60  & 6.36  & 7.53 \\
& 32 & 210.58 & 0.39 & 4.76  & 5.52  & 7.57 \\
& 64 & 171.80 & 0.40 & 5.44  & 6.47  & 6.47 \\
\midrule
\multirow{4}{*}{Ettin-400M}
& 1  & 50.25  & 0.46 & 19.68 & 21.67 & 24.04 \\
& 8  & 101.34 & 0.17 & 9.44  & 15.28 & 20.34 \\
& 32 & 80.67  & 0.22 & 12.58 & 14.27 & 20.52 \\
& 64 & 66.62  & 0.07 & 14.07 & 16.67 & 16.67 \\
\midrule
\multirow{4}{*}{LG4-12B}
& 1  & 1.10 & 0.00 & 917.07 & 1196.54 & 1355.02 \\
& 8  & 1.31 & 0.00 & 717.63 & 963.52  & 1040.36 \\
& 32 & 1.30 & 0.00 & 779.54 & 793.83  & 1043.93 \\
& 64 & 1.20 & 0.00 & 800.36 & 899.83  & 899.83 \\
\bottomrule
\end{tabular}

\vspace{1pt}
\raggedright
\tiny
\textit{Note.} CPS = completions per second. Latency values are in milliseconds.
\end{center}

\end{document}